\pdfoutput=1

\documentclass[11pt]{article}

\usepackage[final]{acl}

\usepackage{times}
\usepackage{latexsym}

\usepackage[T1]{fontenc}

\usepackage[utf8]{inputenc}

\usepackage{microtype}

\usepackage{inconsolata}

\usepackage{graphicx}

\usepackage{tikz}
\usetikzlibrary{shapes, arrows.meta, positioning}
\usepackage{booktabs}
\usepackage{soul}
\usepackage{tabularx}

\graphicspath{{./Images/}}

\title{Label Unification for Cross-Dataset Generalization in Cybersecurity NER}

\author{Maciej Jalocha\thanks{~~Equal contribution.} \\
  {\tt macja@itu.dk} \\\And
    Johan Hausted Schmidt\footnotemark[1] \\
  {\tt jhsc@itu.dk} \\\And
    William Michelsen\footnotemark[1] \\
  {\tt wimi@itu.dk} \\}

\begin{document}
\maketitle

\begin{abstract}
  The field of cybersecurity NER lacks standardized labels, making it challenging to combine datasets. We investigate label unification across four cybersecurity datasets to increase data resource usability. We perform a coarse-grained label unification and conduct pairwise cross-dataset evaluations using RoBERTa models. Qualitative analysis of predictions reveals errors, limitations, and dataset differences. To address unification limitations, we propose alternative architectures including a multi-head model and a graph-based transfer model. Under our selected mapping and training setup, unified training underperformed the corresponding single-dataset baselines on all four development sets. The multi-head model with weight sharing provides no consistent benefits over single-dataset baselines, and the graph-based transfer model shows no consistent benefits. \footnote{Code: \url{https://github.com/maciejjalocha/NLP-Cyber-NER}}.
\end{abstract}

\section{Introduction \& Related Work}
For cybersecurity NER, previous work has noted the scarcity of high-quality, large datasets in English. Due to the cybersecurity domain-specific entity types and vocabulary, it is difficult to make use of general-purpose NER resources \cite{srivastava2023study}. In recent years a handful more datasets have become available \cite{wang2022aptner, wang2020dnrti, alam2022cyner, deka2024attacker}. As of writing this (2025), the cybersecurity domain has no standard set of NER labels, despite previously being proposed as a way of increasing availability of data resources \cite{gao2021review}.  Many of these newer datasets still use different label sets. This makes it difficult for models to combine and use these datasets effectively \cite{gao2021review}. It is, for instance, generally not trivial to re-annotate data for NER in cybersecurity \cite{mouicheti}; no widely applicable, automatic, off-the-shelf annotator exists. Also, cybersecurity often demands familiarity with the domain and manual work for annotation \cite{evangelatos2021named, hanks2022recognizing}. There are other prevalent challenges due to the nature of the cybersecurity domain such as distinctive domain-specific characteristics in cybersecurity reports \cite{marjan2024domain}, which many of the current datasets are based on. Problems such as these are only exacerbated by the domain's data sparsity \cite{liu2022multi}. Various methods have been proposed to remedy the problems caused by, or intimately tied to, the domain's data sparsity. Some have tried to gather as much information about the tokens as possible, such as semantic, morphological, and contextual features, before any classification to better represent rare technical terms and complex nomenclature \cite{marjan2024domain}. Others have similarly tried to integrate many linguistic features to enhance the representation of tokens, by retrieving the most similar words \cite{liu2022multi}.
Previous work has also explored merging labels such that some of the available datasets can be unified to train a more generalizable model \cite{silvestri2022dataset}.

In this work, we investigate the following research question: \textbf{\textit{How does a unification of NER labels across different datasets impact cross-dataset performance in cybersecurity and how can those limitations be overcome with adjustments to models?}}

The main contributions of this investigation are a systematic analysis of cross-dataset performance for recent cybersecurity datasets, enabled by a unification of labels. This is, to the best of our knowledge, unification of a larger set of datasets' labels than has previously been done in cybersecurity NER. The analysis of the cross-dataset predictions helps us understand the differences between the currently available datasets, and shows the viability of such a label unification in practice. Based on what has been identified in the analysis, model-based solutions on these datasets are proposed and evaluated. These contributions can help future research make better use of the available dataset resources in cybersecurity NER.

\section{Methodology}

This section outlines the datasets used in our study, the preprocessing and label unification steps we applied, and the experimental setup used to evaluate cross-dataset performance.

\subsection{Datasets}
Our project uses four different NER cyber-security datasets. APTNER - consisting of 260,134 tokens and 21 different entity types. The dataset consists of reports scraped from various network security companies  \cite{wang2022aptner}. CyNER - with 106,991 tokens and 5 different entity types collected from 60 threat intelligence reports \cite{alam2022cyner}. DNRTI - consisting of 175,220 tokens and 13 entity types. The dataset is comprised of threat intelligence reports from websites of various security companies, government agencies, and GitHub  \cite{wang2020dnrti}. AttackER - consisting of 78,987 tokens and 18 entity types. The dataset is comprised of a mix of reports, articles, and blogs about previous attacks written by cybersecurity experts \cite{deka2024attacker}.  

\subsection{Data Cleaning \& Preprocessing \& Postprocessing}

The datasets we used in this study were originally provided in a variety of formats. To make sure that the different datasets were compatible with our NER models, we converted the datasets to the standard CoNLL 2003 format. CyNER, DNRTI, AttackER used originally BIO and APTNER BIOES.

Since the datasets came with different labels with varying specificity, we standardize their label sets by carrying out a label unification. We decided to go with the labels presented in the CyNER paper. These are "Organization", "System", "Vulnerability", "Malware" and "O". We chose these coarse-grained labels because they generalize well across domains and fit many of the fine-grained labels found in the other datasets. We also decided to exclude the label "Indicator" from the CyNER paper, because the other datasets had no matching labels. In addition, this label can be extracted with regular expressions, rather than contextual understanding \cite{alam2022cyner}.

All labels that could be mapped to one of the four coarse-grained categories were unified accordingly, for all datasets. Labels that could not be meaningfully mapped were discarded and relabelled as O (non-entity). All labels across the datasets that were successfully relabelled can be seen in Table~\ref{tab:label-mapping-appendix}.

\begin{table}[h!]
\centering
\resizebox{\linewidth}{!}{%
\begin{tabular}{llll}
\toprule
\textbf{Dataset} & \textbf{Original Labels} & \textbf{Unified Label} \\
\midrule
\textbf{APTNER} 
& APT, SECTEAM & Organization \\
& OS & System \\
& VULNAME & Vulnerability \\
& MAL & Malware \\

\textbf{DNRTI} 
& HackOrg, SecTeam, org & Organization \\
& Tool & System \\
& Way, exp & Vulnerability \\
& SamFile & Malware \\

\textbf{AttackER} 
& THREAT\_ACTOR, GENERAL\_IDENTITY & Organization \\
& INFRASTRUCTURE, GENERAL\_TOOL, ATTACK\_TOOL & System \\
& VULNERABILITY & Vulnerability \\
& MALWARE & Malware \\

\textbf{CyNER} 
& Indicator & O (removed) \\
\bottomrule
\end{tabular}
}
\caption{Mapping of original entity labels to unified schema used across all four datasets.}
\label{tab:label-mapping-appendix}
\end{table}

Importantly, for the cross-dataset experiment unification occurs after inference. Specifically, a model outputs fine-grained labels' probabilities, and a specific coarse-grained label is chosen by summing these probabilities, and picking the most probable coarse-grained label (Formula~\ref{eq:unification}).
\begin{equation}
    \hat c = \arg\max_{c\in C}\sum_{f\in F_c} P(f|x)
    \label{eq:unification}
\end{equation}
In Formula~\ref{eq:unification}, $C$ is a set of coarse-grained labels, and $F_c$ is a set of the fine-grained labels mapped into a coarse-grained label $c$.
This allowed us to see the models' predictions in fine-grained and coarse-grained label sets.

For the unified model experiment unification occurs before training.

We identified duplicate sentences both within individual datasets (across train, development, and test) and across different datasets. An example can be seen in Table~\ref{tab:duplicate-example}. 
So, from each training dataset the union of the 4 evaluation datasets is subtracted. This applies to the all discussed models.

\begin{table}[h]
\centering
\resizebox{\linewidth}{!}{%
\begin{tabular}{p{2.2cm}p{13cm}}
\toprule
\textbf{Sentence} & \textit{APT32 actors continue to deliver the malicious attachments via spear-phishing emails.} \\
\midrule
\textbf{labels A} & B-Organization O O O O O B-Malware I-Malware O B-Vulnerability O O \\
\textbf{labels B} & B-Organization O O O O O O O O O \\
\bottomrule
\end{tabular}%
}
\caption{Comparison of token-level annotation of the same sentence in two different datasets.}
\label{tab:duplicate-example}
\end{table}

\subsection{Model Architecture}
RoBERTa is used \cite{liu2019robertarobustlyoptimizedbert}, adjusted for NER (Table~\ref{tab:roberta-model-architecture}), with hyperparameters (Table~\ref{tab:roberta-training-params}) from related work \cite{deka2024attacker}.
\begin{table}[h]
\centering
\begin{tabularx}{\linewidth}{lX}
\hline
\textbf{Component} & \textbf{Specification} \\
\hline
Backbone & RoBERTa-base encoder \\
Token Classification Head & Linear layer on top of the encoder hidden states for token labels \\
Input Encoding & Tokenized word-level input with RoBERTa tokenizer; truncation to max length 512 \\
Label Alignment & Only the first subword of each token receives the true label; remaining subwords are masked \\
\hline
\end{tabularx}
\caption{Model architecture for the fine-tuned RoBERTa cross-dataset model}
\label{tab:roberta-model-architecture}
\end{table}

\begin{table}[h]
\centering
\begin{tabularx}{\linewidth}{lX}
\hline
\textbf{Hyperparameter} & \textbf{Value} \\
\hline
Base Model & RoBERTa-base \\
Loss Function & CE (first token only) \\
Optimizer & AdamW \\
Learning Rate & 2e-5 \\
Batch Size & 2 \\
Epochs & 10 \\
Learning Rate Scheduler & Linear decay \\
Warmup Ratio & 0.0 \\
Weight Decay & 0.0 \\
Random Seed & 42 \\
Max Sequence Length & 512 \\
\hline
\end{tabularx}
\caption{Training hyperparameters for the fine-tuned RoBERTa cross-dataset model}
\label{tab:roberta-training-params}
\end{table}

\subsection{Cross-Dataset Evaluation Setup} \label{sec:eval-setup}
We perform pairwise evaluations by training our model on each training dataset and testing it on all other development datasets to assess cross-dataset generalization.

This approach results in a 4x4 matrix where each entry is the span-F1 score. The off-diagonal is the performance of the models trained on one dataset and evaluated on another, and the diagonal entries are from models trained and evaluated on the same dataset.
In addition to span-F1, we also computed recall, precision, and the loose and unlabelled versions of span-F1, which can be found in the results in tables \ref{tab:cross_eval_matrix_precision_roberta}, \ref{tab:cross_eval_matrix_recall_roberta}, \ref{tab:unlabelled_span_f1_cross_dataset_roberta}, and \ref{tab:loose_span_f1_cross_dataset_roberta}. During the analysis, we contrasted the diagonal entry with one set of cross-dataset predictions drawn from a row-adjacent, off-diagonal entry. This was done exactly once for each evaluation dataset. It was also ensured that no off-diagonal model was reused, such that all models were used once for intra-dataset predictions, and once for cross-dataset predictions. We also used Jensen-Shannon divergence (JS-div) to measure similarity between training sets.

\section{Results}

\begin{table}[h!]
\centering
\resizebox{0.9\linewidth}{!}{%
\begin{tabular}{lcccc}
\toprule
\textbf{Train \textbackslash Dev} & \textbf{DNRTI} & \textbf{AttackER} & \textbf{APTNER} & \textbf{CyNER} \\
\midrule
\textbf{DNRTI}    & \textbf{0.63} & 0.30 & 0.21 & 0.16 \\
\textbf{AttackER} & 0.34 & \textbf{0.61} & 0.26 & 0.37 \\
\textbf{APTNER}   & 0.39 & 0.32 & \textbf{0.59} & 0.55 \\
\textbf{CyNER}    & 0.25 & 0.24 & 0.32 & \textbf{0.73} \\
\bottomrule
\end{tabular}
}%
\caption{RoBERTa span-F1 on each development dataset after training on each dataset.}
\label{tab:cross_eval_matrix_roberta}
\end{table}

\begin{table}[h!]
\centering
\resizebox{0.9\linewidth}{!}{%
\begin{tabular}{lcccc}
\toprule
\textbf{Train \textbackslash Dev} & \textbf{DNRTI} & \textbf{AttackER} & \textbf{APTNER} & \textbf{CyNER} \\
\midrule
\textbf{DNRTI}    & \textbf{0.63} & 0.29 & 0.15 & 0.13 \\
\textbf{AttackER} & 0.38 & \textbf{0.61} & 0.20 & 0.31 \\
\textbf{APTNER}   & 0.48 & 0.40 & \textbf{0.51} & 0.58 \\
\textbf{CyNER}    & 0.30 & 0.29 & 0.28 & \textbf{0.76} \\
\bottomrule
\end{tabular}
}%
\caption{RoBERTa precision on each development dataset after training on each dataset.}
\label{tab:cross_eval_matrix_precision_roberta}
\end{table}

\begin{table}[h!]
\centering
\resizebox{0.9\linewidth}{!}{%
\begin{tabular}{lcccc}
\toprule
\textbf{Train \textbackslash Dev} & \textbf{DNRTI} & \textbf{AttackER} & \textbf{APTNER} & \textbf{CyNER} \\
\midrule
\textbf{DNRTI}    & \textbf{0.64} & 0.32 & 0.38 & 0.21 \\
\textbf{AttackER} & 0.32 & \textbf{0.62} & 0.40 & 0.45 \\
\textbf{APTNER}   & 0.33 & 0.27 & \textbf{0.69} & 0.52 \\
\textbf{CyNER}    & 0.22 & 0.21 & 0.37 & \textbf{0.71} \\
\bottomrule
\end{tabular}
}%
\caption{RoBERTa recall on each development dataset after training on each dataset.}
\label{tab:cross_eval_matrix_recall_roberta}
\end{table}

\begin{table}[h!]
\centering
\resizebox{0.9\linewidth}{!}{%
\begin{tabular}{lcccc}
\toprule
\textbf{Train \textbackslash Dev} & \textbf{DNRTI} & \textbf{AttackER} & \textbf{APTNER} & \textbf{CyNER} \\
\midrule
\textbf{DNRTI}    & \textbf{0.71} & 0.43 & 0.37 & 0.44 \\
\textbf{AttackER} & 0.55 & \textbf{0.67} & 0.36 & 0.48 \\
\textbf{APTNER}   & 0.55 & 0.45 & \textbf{0.66} & 0.60 \\
\textbf{CyNER}    & 0.57 & 0.42 & 0.58 & \textbf{0.82} \\
\bottomrule
\end{tabular}
}%
\caption{RoBERTa unlabelled span-F1 on each development dataset after training on each dataset.}
\label{tab:unlabelled_span_f1_cross_dataset_roberta}
\end{table}

\begin{table}[h!]
\centering
\resizebox{0.9\linewidth}{!}{%
\begin{tabular}{lcccc}
\toprule
\textbf{Train \textbackslash Dev} & \textbf{DNRTI} & \textbf{AttackER} & \textbf{APTNER} & \textbf{CyNER} \\
\midrule
\textbf{DNRTI}    & \textbf{0.69} & 0.40 & 0.27 & 0.20 \\
\textbf{AttackER} & 0.44 & \textbf{0.69} & 0.39 & 0.48 \\
\textbf{APTNER}   & 0.44 & 0.41 & \textbf{0.67} & 0.59 \\
\textbf{CyNER}    & 0.29 & 0.38 & 0.35 & \textbf{0.78} \\
\bottomrule
\end{tabular}
}%
\caption{RoBERTa loose span-F1 on each development dataset after training on each dataset.}
\label{tab:loose_span_f1_cross_dataset_roberta}
\end{table}

Table~\ref{tab:cross_eval_matrix_roberta} shows the results of our cross-dataset experiment and, perhaps unsurprisingly, no row-adjacent entry outperforms the relevant diagonal entry in span-F1 score, or precision (table~\ref{tab:cross_eval_matrix_precision_roberta}) or recall (table~\ref{tab:cross_eval_matrix_recall_roberta}). 

There does not seem to be a general pattern between precision and recall. Only when evaluating on APTNER or training on DNRTI recall tends to be higher.

It seems that training on APTNER gives the best cross results. An explanation could be that APTNER is the largest dataset and thus contains the most examples.

There is no single dataset that is easiest to evaluate on.

\section{Analysis}
Cross-dataset predictions give us a way to examine how the datasets differ and where those differences may cause trouble for a unified label set. When a model trained on one dataset fails on another, the failure may point to a disagreement between the two datasets’ labeling approach and definitions, rather than just a model failure. We use this as a proxy to judge whether the unification is practical. 
Below we perform analysis for four pairs of the dataset, followed by the divergence measures between the datasets’ domains.
We sampled the train-dev dataset pairs to analyse in a way to have each dataset once as a dev and once as a train dataset.

\subsection{``AttackER to CyNER`` and ``CyNER to CyNER``}
Discrepancies in predictions here come from primarily three sources: label span length, the treatment of general/generic terms and concepts, and the mapping itself.

AttackER uses much longer label spans than any other dataset in this paper. When its model is applied to CyNER, it overshoots, predicting labels for tokens where CyNER has O, often extending a span across most of a sentence, revealing a difference in annotation approach.

AttackER also labels general concepts. It predicts phrases such as ‘the malware’ and ‘banking Trojan’ as Malware. CyNER primarily reserves its labels for named entities, so it labels ‘Riltok’ as malware, but not ‘the malware’. This is visible upon examination of both datasets’ training data. 

The third source is our mapping. In several cases, the definitions given in the original papers suggested a good match, where the datasets turned out not to agree. For instance, we mapped AttackER’s Infrastructure to System on that basis. Our comparison of predictions reveals that CyNER’s System covers named operating systems and software in a fairly narrow sense, while AttackER’s Infrastructure extends to components such as AccessibilityService. Another example is AttackER’s Threat actor label which shows the same problem. After searching CyNER’s training data to understand its Organization label we see that it does label organizations that have acted maliciously, such as Yingmob, under its Organization label. What separates the two datasets is that CyNER labels cryptonyms and codenames of malicious groups under Malware rather than Organization. 

\subsection{``CyNER to AttackER`` and ``AttackER to AttackER``}
The same trends appear in this direction, with the CyNER model now making the errors. For instance, the CyNER model predicts Malware for entities that AttackER annotates as Threat actor and we mapped to Organization such as MuddyWater, Dukes, and Moses Staff. The mapping of Infrastructure to System also leads to errors as the CyNER model fails to predict spans such as ‘server infrastructure’. Sampling prediction errors in this direction did reveal some annotation errors in AttackER. Tokens such as KillNet and MicroSCADA, which the CyNER model predicts labels for, are labeled as O. This is despite those tokens fitting some of AttackER’s own labels such as Infrastructure or Threat actor.

\subsection{``DNRTI to APTNER`` and ``APTNER to APTNER``}
This pair shows more mapping problems. We mapped DNRTI’s Tool to System, but the tokens it covers are often labeled as malware in APTNER. More importantly, Tool is an example of a label where it turns out no mapping would have worked well, despite the coarse target set borrowed from CyNER.  Reading through DNRTI’s training data, Tool covers named malware families, but also legitimate software such as PowerShell and Word. SamFile is a similar example. It contains both malware, but also generic phrases such as ‘sample’ and ‘documents’, and legitimate Windows Binaries.

There are also cases where despite the mapping and predictions by the DNRTI model being correct, it still fails as APTNER sometimes has no label that covers the relevant tokens. An example is the DNRTI model predicting Way for ‘remote desktop connection’, which maps to Vulnerability. APTNER has no label appropriate for this span.

HackOrg from DNRTI, mapped to Organization, shows two problems. First, the label is built primarily from cryptonyms and codenames which CyNER’s Organization label excludes. Also, DNRTI model errors reveal that APTNER has no label capable of catching such Organizations, since DNRTI often predicts HackOrg for things labeled as O.

\subsection{``APTNER to DNRTI`` and ``DNRTI to DNRTI``}
This direction reveals similar problems that the previous pairs have already shown.

Upon examining the errors we find that labels such as APTNER’s IDTY should have been mapped to Organization. However, this only became clear from examining the predictions since the paper itself does not make the scope of the label obvious.

The APTNER model’s predictions also reveal that the label set of APTNER struggles to cover some coarse labels such as Vulnerability. APTNER has VULNAME, but it is a small vocabulary of roughly 40 forms dominated by ‘exploit’, ‘zero-day’ and ‘vulnerability’, with a few names such as EternalBlue. It does not generalise to phrases such as ‘remote access’ which the model is then punished for on DNRTI.

\subsection{Implications for label unification}

It is clear from this analysis that the datasets differ along several axes at once: How long label spans are, how labels are defined, how individual tokens are treated, etc. Each of these issues will complicate a unification of labels and lead to mixed signals in the merged dataset. For instance, if the datasets disagree on what the coarse labels should cover, it will provide mixed signals to a unified model, since some tokens may be labeled in one part of the dataset, but not in another. Similarly, some datasets caring primarily about named entities while others include generic phrases, which is another source of disagreement within the same tags.

The analysis also reveals a difficult trade-off. Even with CyNER’s labels as a foundation for a coarse label set, mapping is difficult. Many labels, if mapped, risk bringing conflicting content into the coarse tags. However, if one wants to completely avoid this a significant portion of labels have to be dropped from the mapping entirely.

Our scores in Table~\ref{tab:cross_eval_matrix_roberta} are meant to reflect what a practical, coarse unification produces. The mapping could be improved, but doing so required reading through raw datasets and searching for keywords, to work out what the various labels cover. If such manual effort is a precondition for adding each dataset, we take it as evidence that the field needs either more standardized labels, or at least better documented annotation schemes. A few extra sentences or examples defining each label in the original papers would have made our unification considerably easier.

\subsection{Language Metrics}

\begin{table}[h]
\centering
\resizebox{0.9\linewidth}{!}{%
\begin{tabular}{lcccc}
\toprule
\textbf{Datasets} & \textbf{DNRTI} & \textbf{AttackER} & \textbf{APTNER} & \textbf{CyNER} \\
\midrule
\textbf{DNRTI} & 0 & 0.23 & 0.04 & 0.05 \\
\textbf{AttackER} & 0.23 & 0 & 0.24 & 0.19 \\
\textbf{APTNER} & 0.04 & 0.24 & 0 & 0.07 \\
\textbf{CyNER} & 0.05 & 0.19 & 0.07 & 0 \\
\bottomrule
\end{tabular}%
}
\caption{JS-div between training datasets for distributions over span length of entities. }
\label{tab:spanlength-jsdiv}
\end{table} 

Information-theoretic measures like JS-div estimate differences between probability distributions \cite{kashyap2020domain}. In NLP, these are often based on relative frequencies \cite{lu2021diverging}. 

We notice two things. Firstly, our span-length divergence estimates in Table~\ref{tab:spanlength-jsdiv} show AttackER stands out, aligning with observations in the analysis.
Secondly, divergence in words distribution, and other, seems to be unrelated to RoBERTa performance (Table~\ref{tab:roberta_js_correlation}).

\begin{table}[h!]
\centering
\resizebox{0.9\linewidth}{!}{%
\begin{tabular}{lrrrr}
\toprule
\textbf{Distribution} & \textbf{RoBERTa} & p-value \\
\midrule
Words & -0.02 & 0.95 \\
POS labels & -0.23 & 0.48 \\
Entity span lengths & 0.00 & 0.99 \\
Entity span-based counts & 0.13 & 0.68 \\
\bottomrule
\end{tabular}
}%
\caption{Pearson correlations between JS divergence and cross-dataset span-F1 over off-diagonal dataset pairs.}
\label{tab:roberta_js_correlation}
\end{table}

\subsection{Unified-datasets model}
We hypothesise that if we train the model on all four unified datasets, that the scores on the individual evaluation datasets should be higher than a model trained only on the individual datasets. Table~\ref{tab:bigmodel_vs_original} shows the difference in span-F1 performance between predicting a development dataset with a model trained on all four datasets compared to a model only trained on a single dataset.
\begin{table}[h!]
\centering
\resizebox{0.9\linewidth}{!}{%
\begin{tabular}{lcccc}
\toprule
\textbf{Dataset} & \textbf{Unified} & \textbf{Train-Only} \\
\midrule
Unified & 0.53 & - \\
DNRTI    & 0.52 & 0.63 \\
AttackER & 0.51 & 0.61 \\
APTNER   & 0.49 & 0.59 \\
CyNER    & 0.67 & 0.73 \\
\bottomrule
\end{tabular}
}%
\caption{Span-F1 on development datasets for a model trained on unified dataset or a corresponding train dataset.}
\label{tab:bigmodel_vs_original}
\end{table}

Opposite to the expectation, training on a unified dataset leads to a decrease in performance across the board. It suggests that the annotation conflicts introduced, and surfaced in the analysis, limit benefits of more data, although other training effects may contribute.

\section{Addressing Unification Limitations}
\subsection{Multi-head Model}

\begin{table}[h!]
\centering
\begin{tabular}{lccc}
\toprule
\textbf{Dataset} & \textbf{Reference} & \textbf{Multi-head} \\
\midrule
DNRTI    & 0.62 & 0.61 \\
AttackER & 0.43 & 0.42 \\
APTNER   & 0.59 & 0.60 \\
CyNER    & 0.77 & 0.80 \\
\bottomrule
\end{tabular}
\caption{Span-F1 per dataset for RoBERTa multi-head vs single-dataset fine-grained reference.}
\label{tab:roberta_multihead_vs_reference}
\end{table}

Some of the identified error trends from the analysis likely explain the disappointing results in Table~\ref{tab:bigmodel_vs_original}. Here, we present the performance of a model that uses all available datasets without applying a label unification to the label sets. It is a multi-headed model where each head predicts labels from a dataset's fine-grained label set. While the weights associated with the output heads are individualised, those associated with the transformers blocks may be shared. Such a model architecture could improve performance as the tasks demanded by the datasets are similar

A reference model was also trained for each dataset, which do not share any weights, but do use the same label sets as the multi-head model, for a baseline point of comparison.

Looking at Table~\ref{tab:roberta_multihead_vs_reference}, there is no clear difference the multi-head and reference models.  

It is interesting that for 3 out of 4 of the datasets, span-f1 scores for the multi-head variant and the unification-model (Table~\ref{tab:roberta_multihead_vs_reference}, Table~\ref{tab:bigmodel_vs_original}) are similar, despite multi-head model having more classes to classify to. An exception is AttackER. Upon inspection of its labels, we hypothesise that the fine-grained AttackER's labels retained for the unified model (Table~\ref{tab:label-mapping-appendix}) were easier to distinguish between each other than the dropped ones, thus unified model had easier task than the multi-head model.

\subsection{LST-NER}

\begin{table}[h!]
\small
\centering
\begin{tabular}{|l|c|c|}
\hline
 \textbf{Dataset} & \textbf{Baseline (RoBERTa)} & \textbf{LST-NER (ours)} \\
\hline
DNRTI     & 0.72          & 0.70$^{*}$ \\
AttackER & 0.44          & 0.43       \\
APTNER    & 0.51          & 0.48$^{*}$ \\
CyNER     & 0.76          & 0.81$^{*}$ \\
\hline
\end{tabular}
\caption{Span-F1 comparing our RoBERTa baseline and the graph-matching model (LST-NER). $^{*}$ indicates a statistically significant difference (paired bootstrap test, $p < 0.05$); AttackER shows no significant difference ($p = 0.35$).}
\label{tab:spanf1-graph}
\end{table}

The last model we evaluated is based on the graph matching from the paper "Cross-domain Named Entity Recognition via Graph Matching" \cite{zheng2024crossdomain}. We reimplement the paper's core mechanism estimating label representations from a source-domain model, constructing a graph over label relationship, and refining these representations with a graph neural network. Lastly, we align the source and target label graph using the proposed Gromov-Wasserstein distance. Unlike our label unification process, this architecture takes a different approach by preserving the original label sets and leveraging structural knowledge from a rich-resource domain. 

The key insight in the paper is that a pre-trained NER model from a general domain can provide structural knowledge about entity relationships even if the labels do not match those of the target domain via graph matching. In theory, this should allow for the transfer of knowledge from the resourceful dataset to the more sparse dataset without requiring label correspondence. While the original paper uses BERT-base-case, we swapped the backbone to instead use Roberta \cite{liu2019robertarobustlyoptimizedbert} to match the model used through the paper. In addition, we also compared it to a RoBERTa baseline for a fair comparison model without the addition of graphs proposed in the paper. Table~\ref{tab:spanf1-graph} demonstrates the results on each evaluation dataset using a paired bootstrap test following standard practice in NLP \cite{koehn-2004-statistical, berg-kirkpatrick-etal-2012-empirical}. The results are mixed. The baseline significantly outperforms the graph-matching model on APTNER and DNRTI ($p < 0.05$), whereas the graph model significantly outperforms the baseline on CyNER ($p < 0.05$). The two remain statistically insignificant on AttackER. This suggests that the graph based method is dataset dependent rather than having a consistent effect across our four cybersecurity NER datasets. 

Hyperparameters for the models can be found in Table~\ref{tab:lst-ner-model-params}.

\begin{table}[h]
\centering
\begin{tabularx}{\linewidth}{lX}
\hline
\textbf{Hyperparameter} & \textbf{Value} \\
\hline
Learning rate & 5e-5\\
Batch Size & 16 \\
Epochs & 3 \\
Max Length & 128 \\
Temp & 4 \\
Edge Threshold & 1.5 \\
gwd\_lambda & 0.01 \\

\hline
\end{tabularx}
\caption{Training hyperparameters for RoBERTa and LST-NER (only the first four applies to RoBERTa, whereas all hyperparameters apply LST-NER).}
\label{tab:lst-ner-model-params}
\end{table}

\section{Conclusion}
We investigated the effects of label unification across cybersecurity NER datasets, by evaluating cross-dataset performance. While unification enabled unified training datasets, it introduced errors due to annotation differences and label mismatches. Unified models underperformed compared to single-dataset baselines.

Multi-head models, with an exception, showed comparable performance while preserving original label sets, and LST-NER, despite its complexity, offered no clear advantage over RoBERTa. Overall, our results suggest that simple label merging is not sufficient for robust generalization, and future work should explore more targeted domain adaptation strategies.

\newpage
\nocite{*}

\newpage

\section*{Conflict of interest}
Authors state no conflict of interest.

\appendix
\section*{Appendix}
\label{sec:appendix}
\section{Limitations}
We did not perform extensive hyperparameter tuning. Consequently, optimal configurations may yield different performance results. Additionally, the label unification process was conducted without external verification from cybersecurity domain experts, which may have resulted in suboptimal mappings. Finally, while we utilized baseline architectures to test the unification, more complex models might be required to fully leverage the unified data.

Furthermore, our experiments do not isolate fully the effects of annotation conflicts from those of the model or the training parameters.

\section{Data Cleaning}
This appendix summarises the deterministic cleaning rules applied to the original datasets
before label–unification. 

\subsection{\texttt{APTner (CoNLL)}}
\begin{itemize}
    \setlength\itemsep{0pt}
    \setlength\parsep{0pt}
    \setlength\topsep{0pt}
  \item Sentences are forced to break after the literal line \texttt{. O}, reproducing
        the sentence count reported in the original paper. Original file breaks are not considered.
  \item A line that contains only \texttt{O} is discarded.
  \item Any other single‑token line is retained with its label changed to \texttt{O}.
  \item A line with \(\ge 3\) whitespace‑separated fields is treated as corrupt; the first
        token is preserved and labelled \texttt{O}.
  \item A two‑token line whose second field is \emph{not} in the official
        \texttt{labels} set is kept, but its label is replaced by
        \texttt{O}.
\end{itemize}

\subsection{\texttt{DNRTI (CoNLL)}}
\begin{itemize}
\setlength\itemsep{0pt}
  \setlength\parsep{0pt}
  \setlength\topsep{0pt}
  \item A line that contains only \texttt{O} is discarded.
  \item Any other single‑token line is retained with its label changed to \texttt{O}.
  \item For A line with \(\ge 3\) whitespace‑separated fields the first
        token is preserved and labelled \texttt{O}.
\end{itemize}

\subsection{\texttt{Attacker (JSON)}}
\begin{itemize}
\setlength\itemsep{0pt}
  \setlength\parsep{0pt}
  \setlength\topsep{0pt}
  \item Tokens that are an explicit space character (i.e.\ the string \verb|" "|) are
        skipped. \emph{Note:} this check occurs inside the label‑unification function, so no
        separate \textit{cleaned} intermediate file is produced.
\end{itemize}
\vspace{2pt}
All these operations are reproducible using the cleaning functions for DNRTI \& Aptner given their raw files. The cleaned files are produced.

\section{Acknowledgements}
We thank Dr. Rob van der Goot (IT University of Copenhagen) for his supervision and guidance throughout the writing of this paper.

\section*{Disclosure of Chatbot Use}

In accordance with the ACL 2023 policy on the use of generative AI tools, we disclose the following usage of a chatbot (OpenAI's ChatGPT) in the development of this project:

\textbf{Implementation Support} \\
ChatGPT was used to assist in the implementation of an LST-NER model. The original research paper (to the best of our knowledge) did not provide a public code repository. The chatbot was used to generate initial code snippets and to help debug issues encountered during development.

ChatGPT also assisted with an internal review of the paper.

All code was critically reviewed, tested, and adapted by the authors to ensure alignment with the methodology described in the paper.

The final results and interpretations presented in this work are entirely the responsibility of the authors.

\section*{Training details for the unified and multi-head models}

During training the dataset are sampled, with replacement, proportional to each dataset's sentence count. Sentences from the same dataset compose a batch. Batches from the same datasets are sampled consecutively, in a cycle.

\subsection*{Hardware and accounting unit}

\begin{table}[h!]
\centering
\begin{tabular}{lcccc}
\toprule
\textbf{Experiment} & \textbf{Training config} \\
\midrule
Cross-dataset &  \textbf{1.09 GCD-h} \\
unified &  \textbf{1.15 GCD-h} \\
Multi-head & \textbf{1.18 GCD-h} \\
\bottomrule
\end{tabular}
\caption{Compute-only RoBERTa fine-tuning costs measured on LUMI. Each experiment run on one GCD (Graphic Compute Die) of one AMD MI250x node.}
\label{tab:lumi_gpu_hours}
\end{table}

\begin{table}[h!]
\centering
\begin{tabular}{lcc}
\toprule
\textbf{Dataset} & \textbf{GPU} & \textbf{GPU-hours} \\
\midrule
AttackER & RTX 2070 & 0.37 \\
APTNER   & A30      & 0.93 \\
DNRTI    & RTX 2070 & 0.68 \\
CyNER    & RTX 6000 & 0.24 \\
\midrule
\textbf{Total} & --- & \textbf{2.22} \\
\bottomrule
\end{tabular}
\caption{GPU time for the graph-model experiment. Runs used different GPU models; GPU-hours represent elapsed GPU time, not normalized compute.}
\label{tab:graph-gpu-hours}
\end{table}

\section*{Code and Data}
Code, datasets, and instructions for reproducing the experiments are available at \url{https://github.com/maciejjalocha/NLP-Cyber-NER}

\bibliography{custom}
\end{document}